\documentclass[runningheads]{llncs}
\usepackage{amsmath}
\usepackage{amsfonts}
\usepackage{graphicx}
\usepackage{hyperref}
\usepackage{tabularx}

\hypersetup{
    colorlinks=true,
    citecolor=blue,
    linkcolor=blue,
    filecolor=magenta,      
    urlcolor=cyan}

\begin{document}
\title{Inverse Consistency by Construction for Multistep Deep Registration}

\author{Hastings Greer\inst{1} \and
	Lin Tian\inst{1} \and
    Francois-Xavier Vialard \inst{2} \and
    Roland Kwitt \inst{3} \and 
    Sylvain Bouix \inst{4} \and
    Raul San Jose Estepar \inst{5} \and
    Richard Rushmore \inst{6} \and
	Marc Niethammer\inst{1}}
%
%
\institute{University of North Carolina at Chapel Hill \and
University Paris-Est \and
University of Salzburg, Austria \and
ÉTS Montréal \and
Brigham and Women's Hospital \and
Boston University
}
\maketitle              
\begin{abstract}
	Inverse consistency is a desirable property for image registration. We propose a simple technique to make a neural registration network inverse consistent by construction, as a consequence of its structure, as long as it parameterizes its output transform by a Lie group. We extend this technique to multi-step 
	neural registration by composing many such networks in a way that preserves inverse consistency. This multi-step approach also allows for inverse-consistent coarse to fine registration. We evaluate our technique on synthetic 2-D data and four 3-D medical image registration tasks and obtain excellent registration accuracy while assuring inverse consistency.
	\keywords{Registration  \and Deep Learning}
\end{abstract}
\vspace{-2em}
\begin{figure}
	\centering{
		\includegraphics[width=.16\textwidth]{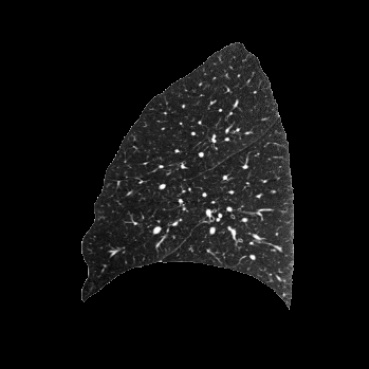}
		\includegraphics[width=.16\textwidth]{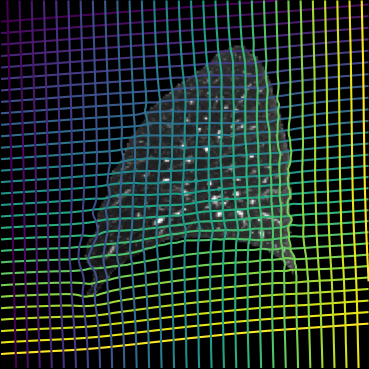}
		\includegraphics[width=.16\textwidth]{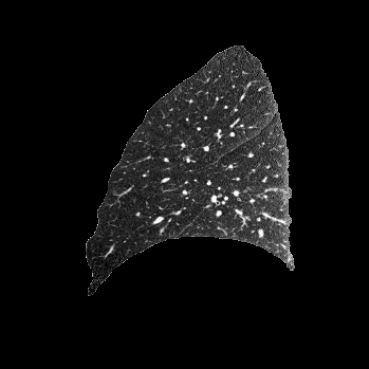}
		\includegraphics[width=.16\textwidth]{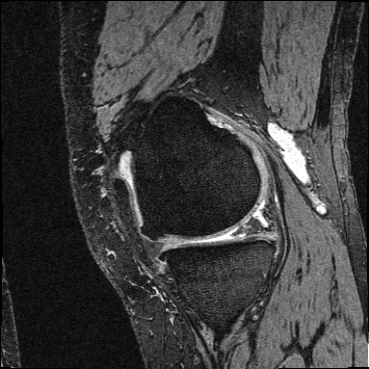}
		\includegraphics[width=.16\textwidth]{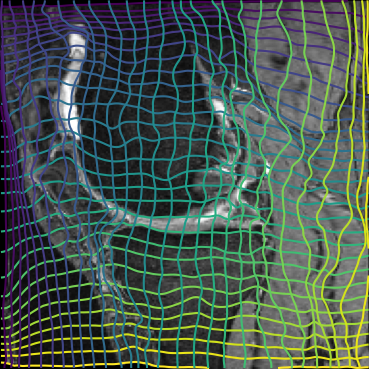}
		\includegraphics[width=.16\textwidth]{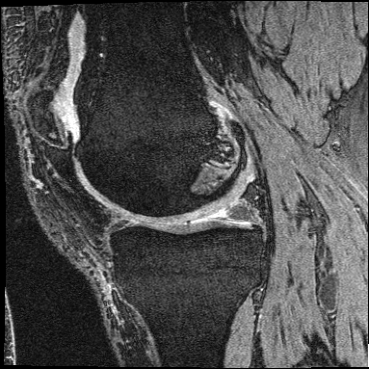}}
	\caption{Cases registered by \texttt{ConstrICON} from DirLab and OAI}
	\label{fig:medical_sample}
	\vspace{-2em}
\end{figure}

\section{Introduction}
\label{sec:introduction}

Image registration, or finding the correspondence between a pair of images, is
a fundamental task in medical image computing. One desirable property for
registration algorithms is inverse consistency -- the property that the
transform found registering image A onto image B, composed with the transform
found by registering image B onto image A, yields the identity map. Inverse
consistency is useful for several reasons. Practically, it is convenient to
have a single transform and its inverse associating two images instead of two
transforms of unknown relationship. For within-subject registration, inverse
consistency is often a natural assumption as long as images are consistent
with each other, e.g., did not undergo surgical removal of tissue. For time
series analysis, inverse consistency prevents
bias~\cite{REUTER20101181}. We propose a novel deep network structure
that registers images in multiple steps in a way that is
\emph{inverse-consistent by construction}. Our approach is flexible and allows
different transform types for different steps.

\section{Related Work}
\label{sec:related_work}

Inverse consistency in deep image registration approaches is commonly promoted 
via a penalty~\cite{greer2021icon,shen2019networks,jun2018ICNet,Nazib2021Cnn}
on the inverse consistency error. Extensive work also exists on
optimization-based \emph{exactly} inverse consistent image registration. For example, by using a symmetric image similarity measure and an inverse consistency loss on the transformations~\cite{christensen2001consistent} or by performing robust inverse consistent rigid registrations with respect to a middle space~\cite{REUTER20101181}.
ANTs SyN~\cite{avants2008symmetric} is an
approach to inverse consistent deformable registration, but by default is part
of a multi-step affine then SyN pipeline which is not as a whole inverse
consistent. 


\vskip0.5ex
Mok \emph{et al.}~\cite{Mok_2020_CVPR} introduce a deep-learning framework that is
exactly inverse consistent. They
take advantage of the fact that a stationary velocity field (SVF) transform representation allows for
fast inversion of a transform by integrating the negated velocity
field. Thus, by calling their network twice, the second time with the inputs
reversed, they can construct a transform $\Phi^{AB} = \exp(N_\theta[I^A, I^B])
	\circ \exp(-N_\theta[I^B, I^A])$. This registration network is
inverse-consistent by construction, but only supports one step. \emph{Our approach will provide a general inverse consistent multi-step framework.}

\vskip0.5ex
Iglesias \emph{et al.}~\cite{iglesias2023easyreg} introduce a two-step deep registration framework for brain registration that is
inverse consistent by construction. First, they
independently segment each image with a U-Net into 97 anatomical regions. The centroids of
these regions and the corresponding regions of an atlas are then used to obtain an affine transformation to the atlas. This is inverse consistent.
Second, each brain image is resampled to the atlas
space followed by an SVF-based transformation, where the velocity field is obtained by two calls to their velocity field network: $\exp(N_\theta[I^A, I^B] - N_\theta[I^B, I^A])$. This symmetrization retains inverse consistency and is conceptually similar to our approach. \emph{However, their approach, unlike ours,  does not directly extend to $N$ steps and is not trained end to end.}
\vskip0.5ex
There is extensive literature on deep multi-step approaches. The core idea
is to conduct the registration in multiple steps with the warped image
produced by the previous step being the input to the latter step. Thus, the
original input image pairs can be registered progressively. AVSM
\cite{shen2019networks} achieves this by reusing the same neural network at each
step. Other works in the literature~\cite{mok2020large,greer2021icon,lin2022GradICON}
setup different neural networks at each step. In addition, these steps are often conducted in a coarse-to-fine manner. Namely, the neural network at the current
step registers the input images at a coarse resolution, interpolates the
output deformation field to a finer resolution, composes the interpolated
deformation field with the composed transformation from previous steps, warps
the moving image at the finer resolution, and passes the warped image and target
image to the neural network at next step.
Greer \emph{et al.} and Tian \emph{et al.}~\cite{greer2021icon,lin2022GradICON} define an abstract \texttt{TwoStep} operator to represent the process described above. However, this \texttt{TwoStep} operation does not guarantee inverse consistency between
the \emph{composed} forward transformation and the \emph{composed} backward
transformation.
\emph{To address this issue, we propose a novel operator for multi-step registration to obtain inverse consistent registration by construction.}








\vskip2ex
\noindent
\textbf{Definitions \& Notation.} We use subscripted capital letters, e.g., $N_\theta$, to represent neural
networks that return arrays of numbers, and capital Greek letters $\Phi, \Psi,
	\text{and}~ \Xi$ to represent registration neural networks, i.e., neural networks
that return transforms. A transform is a function $\mathbb{R}^D \rightarrow
	\mathbb{R}^D$ with $D$ denoting the dimension of the images we are registering.
$N_\theta^{AB}$ is shorthand for $N_\theta$ called on the images $I^A$ and
$I^B$, likewise $\Phi^{AB}$ is shorthand for $\Phi[I^A, I^B]$. A deep
registration network outputs a transform such that $I^A \circ \Xi^{AB} \sim
	I^B$. For a Lie group $G$ and associated algebra $\mathfrak{g}$, $\exp$ is the
(Lie-)exponential map from $\mathfrak{g} \rightarrow G$ \cite{lie1880meme,eade2013lie}.

\section{Lie-group based inverse consistent registration}
\label{sec:lie_group_ic_reg}

To design a registration algorithm, one must pick a class of transforms that
the algorithm will return. Many types of transforms that are useful for
practical medical registration problems happen to also be Lie groups. We
describe a procedure for designing a neural network that outputs a member of a
specified Lie group in an inverse consistent manner and provide several
examples.

\vskip0.5ex
Recall that a Lie group $G$ is always associated with a Lie algebra $\mathfrak{g}$.
Create a neural network $N_\theta$ (of arbitrary design) with two input images and an output that can be considered an element of $\mathfrak{g}$.

A registration network $\Phi$ defined to act as follows on two images
\begin{equation}
	\Phi[I^A, I^B] := \exp(g(I^A,I^B))\,,\quad g(I^A,I^B) := N_\theta[I^A, I^B] - N_\theta[I^B, I^A]
\end{equation}
is inverse consistent, because $g(I^A,I^B) = -g(I^B,I^A)$ by construction. We explore how this applies to several Lie groups.

\vskip0.5ex
\noindent
\textbf{Rigid registration.}
The Lie algebra of rigid rotations is skew-symmetric matrices. $N_\theta$ outputs a skew-symmetric matrix $R$ and a vector $t$, so that
\begin{equation}
	N_\theta^{AB} = \begin{bmatrix}
		R & t \\ 0 &1
	\end{bmatrix}, \Phi_{(\text{rigid})}[I^A, I^B](x) := \exp(N_\theta[I^A, I^B] - N_\theta[I^B, I^A]) \begin{bmatrix}
		x \\ 1
	\end{bmatrix}\,,
\end{equation}
where $\Phi_{(\text{rigid})}$ will output a rigid transformation in an inverse consistent manner. Here, the exponential map is just the matrix exponent. 

\vskip0.5ex
\noindent
\textbf{Affine registration.}
Relaxing $R$ to be an arbitrary matrix instead of a skew-symmetric matrix in the
above construction produces a network that performs inverse consistent affine
registration.

\vskip0.5ex
\noindent
\textbf{Nonparametric vector field registration.}
In the case of the group of diffeomorphisms, the corresponding Lie algebra\footnote{Although in infinite dimensions, the name Lie algebra does not apply, in our case we only need the notions of the exponential map and tangent space at identity to preserve the inverse consistency property.} is the space of vector
fields. If $N_\theta$ outputs a vector field, implemented as a grid of vectors
which are linearly interpolated, then, by using scaling and
squaring~\cite{arsigny2006log,balakrishnan2019voxelmorph} to implement the Lie exponent, we
have
\begin{equation}
	\Phi_{(\text{svf})}[I^A, I^B](x) := \exp(N_\theta[I^A, I^B] - N_\theta[I^B, I^A])(x)\,,
\end{equation}
which is an inverse consistent nonparametric registration network. This is equivalent to the standard SVF technique for image registration, with a velocity field represented as a grid of vectors equal to $N_\theta[I^A, I^B] - N_\theta[I^B, I^A]$.

\vskip0.5ex
\noindent
\textbf{MLP registration.}
An ongoing research question is how to represent the output transform as a multi-layer perceptron (MLP) applied to coordinates. One approach is to reshape the vector of outputs
of a ConvNet so that the vector represents the weight matrices defining an MLP
(with $D$ inputs and $D$ outputs). This MLP is then a member of the Lie algebra of
vector-valued functions, and the exponential map to the group of
diffeomorphisms can be computed by solving the following differential equation to $t=1$ using an integrator such as fourth-order Runge-Kutta. Again, by defining the velocity field to flip signs when the input image order is flipped, we obtain an inverse consistent transformation:
\begin{multline}
	\label{eqn:svf_def}
	v(z) = N_\theta^{AB}(z) - N_\theta^{BA}(z)\,,
	\frac{\partial}{\partial t} \Phi^{AB}(x, t) = v(\Phi^{AB}(x, t))\,, \\ \Phi^{AB}(x, 0) =
	x,~~\Phi^{AB}(x) = \Phi^{AB}(x, 1)\,.
\end{multline}
\section{Multi-step registration}
\label{sec:multi_step}
The standard approach to composing two registration networks is to register
the moving image to the fixed image, warp the moving image and then register
the warped moving image to the fixed image again and compose the transforms.
This is formalized in \cite{greer2021icon,lin2022GradICON} as \texttt{TwoStep}, i.e.,
\begin{equation}
	\texttt{TwoStep}\{\Phi, \Psi\} := \Phi[I^A, I^B] \circ \Psi[I^A \circ \Phi[I^A, I^B], I^B]\,.
\end{equation}
\emph{Unfortunately, \texttt{TwoStep}~\cite{greer2021icon,lin2022GradICON} is not always inverse consistent even  with inverse consistent arguments}. \emph{First}, although $\Psi[\tilde{I^A}, I^B]$ is the inverse of $\Psi[I^B, \tilde{I^A}]$, it does not necessarily have any relationship with $\Psi[\tilde{I^B}, I^A]$ which is the term that appears when swapping the inputs to \texttt{TwoStep}. \emph{Second}, composing \\ $\texttt{TwoStep}[\Phi, \Psi](I^A, I^B) \circ \texttt{TwoStep}[\Phi, \Psi](I^B,
	I^A)$, results in $\Phi \circ \Psi \circ \Phi^{-1} \circ \sim\Psi^{-1}$. The
inverses are interleaved so that even if they were exact, they
can not cancel.

\vskip0.5ex
\emph{Our contribution} is an operator, \texttt{TwoStepConsistent}, that is inverse consistent if its components are inverse consistent. We assume that our component networks $\Phi$ and $\Psi$ are inverse
consistent, and that $\Phi$ returns a transform that we can explicitly find
the square root of, such that $\sqrt{\Phi^{AB}}\circ\sqrt{\Phi^{AB}} =
	\Phi^{AB}$. Note that for transforms defined by $\Phi^{AB}=\exp(g)$ , $  \sqrt{\Phi^{AB}}=\exp(g/2)$. Since each network is inverse consistent, we have access to the
inverses of the transforms they return. We begin with the relationship that $\Phi$ will be trained to fulfill
\begin{equation}
	I^{A} \circ \Phi[I^A, I^B] \sim I^B\,,~\hat{I}^A := I^{A} \circ \sqrt{\Phi[I^A, I^B]} \sim \hat{I}^B := I^B \circ \sqrt{\Phi[I^B, I^A]}\,,
\end{equation}
and apply $\Psi$ to register $\hat{I}^A$ and $\hat{I}^B$
\begin{eqnarray}
	I^{A} &\circ& \sqrt{\Phi[I^A, I^B]} \circ \Psi[\hat{I}^A, \hat{I}^B] \simeq I^B \circ \sqrt{\Phi[I^B, I^A]}\,,\\
	I^{A} &\circ& \sqrt{\Phi[I^A, I^B]} \circ \Psi[\hat{I}^A, \hat{I}^B] \circ
	\sqrt{\Phi[I^A, I^B]}\simeq I^B\,. \label{eqn:tsc_pre}
\end{eqnarray}
We isolate the transform in the left half of Eq.~\eqref{eqn:tsc_pre} as our new operator, i.e.,
\begin{equation}
	\label{eqn:twostepdef}
	\texttt{TwoStepConsistent}\{\Phi, \Psi\}[I^A, I^B]:= \sqrt{\Phi[I^A, I^B]} \circ \Psi[\hat{I}^A, \hat{I}^B] \circ \sqrt{\Phi[I^A, I^B]}\,.
\end{equation}
In fact, we can verify that
\begin{align}
	\texttt{TwoStepConsistent} & \{\Phi, \Psi\}[I^A, I^B] \circ \texttt{TwoStepConsistent}\{\Phi, \Psi\}[I^B, I^A]                                                \\
	                           & = \sqrt{\Phi} \circ \Psi \circ \sqrt{\Phi} \circ \sqrt{\Phi^{-1}} \circ \Psi^{-1} \circ \sqrt{\Phi^{-1}}= \textrm{id} \enspace .
\end{align}

\noindent
\emph{Notably, this procedure extends to $N$-Step registration.} With the operator of Eq.~\eqref{eqn:twostepdef}, a registration network composed from an arbitrary number of steps may be made inverse consistent. This is because $\texttt{TwoStepConsistent}\{\cdot,\cdot\}$ is a valid \emph{second} argument to
\texttt{TwoStepConsistent}. For instance, a three-step network can be constructed as
$	\texttt{TwoStepConsistent}\{\Phi, \texttt{TwoStepConsistent}\{\Psi, \Xi\}\}$.


\section{Synthetic experiments}
\label{sec:synthetic_experiments}

\begin{figure}
	\centering
	\includegraphics[width=1.0\textwidth]{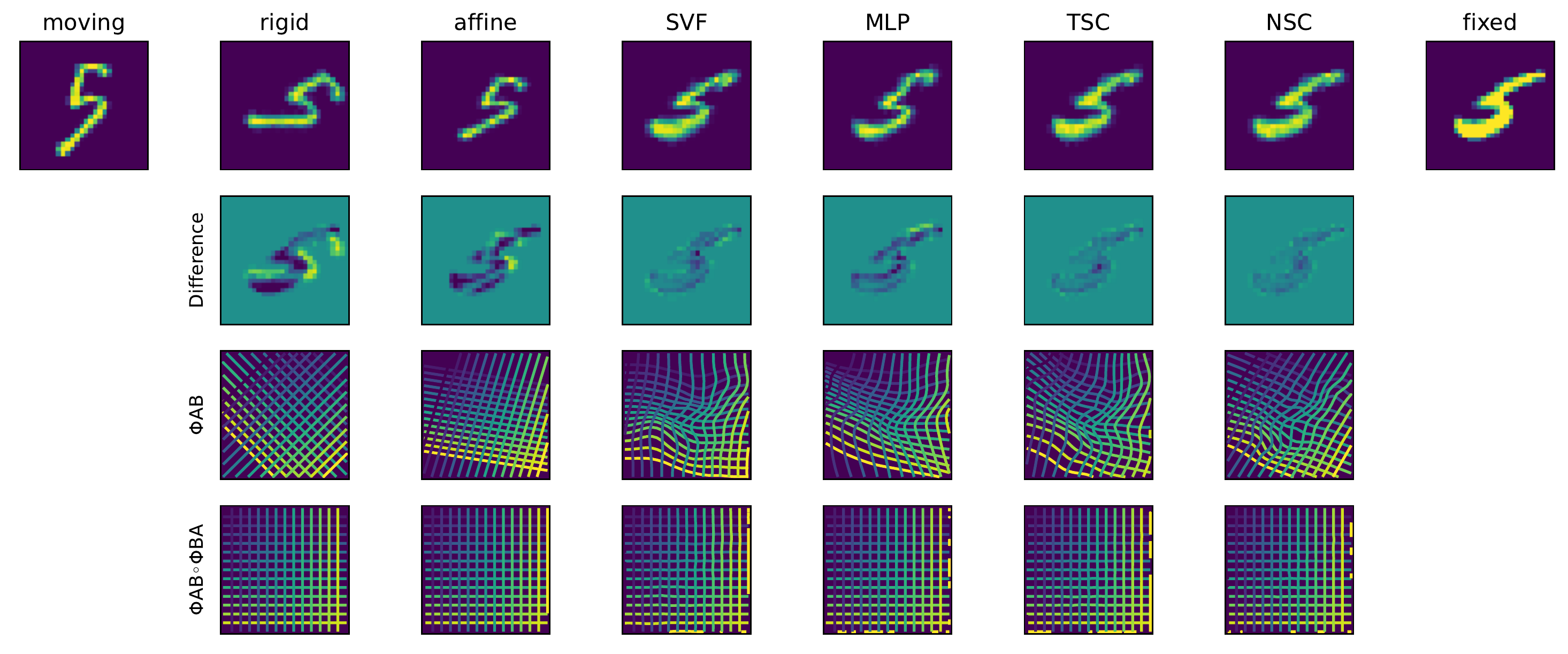}
	\caption{We train single-step rigid, affine, vector field parameterized SVF, and neural deformation field (MLP) networks, as well as a two-step registration network (TSC) composed of a neural deformation field step followed by a vector field parameterized SVF step and a 4 step network (NSC) composed of two affine steps and two SVF steps.  We observe excellent registration results indicated by the small differences (second row) after applying the estimated transformation $\Phi^{AB}$ (third row). Composing with the inverse produces results very close to the identity map (last row) as desired.  
 }
	\label{fig:zoo}
	\vspace{-1em}
\end{figure}

\textbf{Inverse consistent rigid, affine, nonparametric, and MLP registration.} We train networks on MNIST 5s using the methods in Secs.~\ref{sec:lie_group_ic_reg} and~\ref{sec:multi_step}, demonstrating
that the resulting networks are inverse-consistent. Our \texttt{TwoStepConsistent} \texttt{(TSC)} operator can be used on any combination of the networks defined in Sec.~\ref{sec:lie_group_ic_reg}. For demonstrations, we
join an MLP registration network to a vector field registration network, and join two affine networks to two vector field networks. Fig. \ref{fig:zoo} shows successful inverse-consistent sample registrations.

\vskip0.5ex
\noindent
\textbf{Affine registration convergence.}
In addition to being inverse consistent, our method accelerates convergence
and stability of affine registration, compared to directly predicting the
matrix of an affine transform. Here, we disentangle whether this happens for
any approach that parameterizes an affine transform by taking the exponent of
a matrix, or whether this acceleration is unique to our inverse consistent
method. We also claim that multi-step registration is important for registration
accuracy and convergence time and that an \emph{inverse consistent} multi-step
operator, \texttt{TwoStepConsistent}, is thus beneficial.

To justify these claims, we investigate training for affine registration on
the synthetic \emph{Hollow Triangles and Circles} dataset
from~\cite{lin2022GradICON} while varying the method used to obtain a matrix
from the registration network and the type of multi-step registration used. To
obtain an affine matrix, we either directly use the neural network output
$N_\theta^{AB}$, use $\exp(N_\theta^{AB})$, or, as suggested in
Sec.~\ref{sec:lie_group_ic_reg}, use $\exp(N_\theta^{AB} - N_\theta^{BA})$. We
either register in one step, use the \texttt{TwoStep} operator
from~\cite{greer2021icon,lin2022GradICON}, or use our new
\texttt{TwoStepConsistent} operator. This results in 9 training
configurations, which we run 65 times each.

\begin{figure}[t!]
	\centering
	\includegraphics[width=0.32\textwidth]{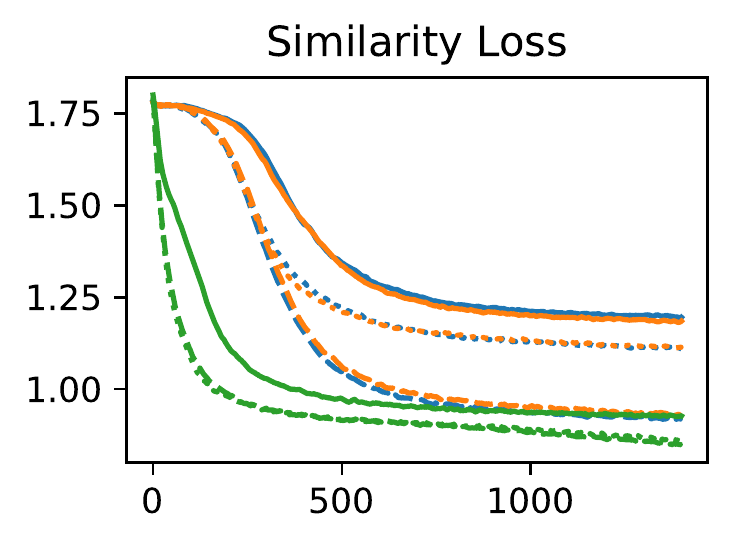}\hfill
	\includegraphics[width=0.32\textwidth]{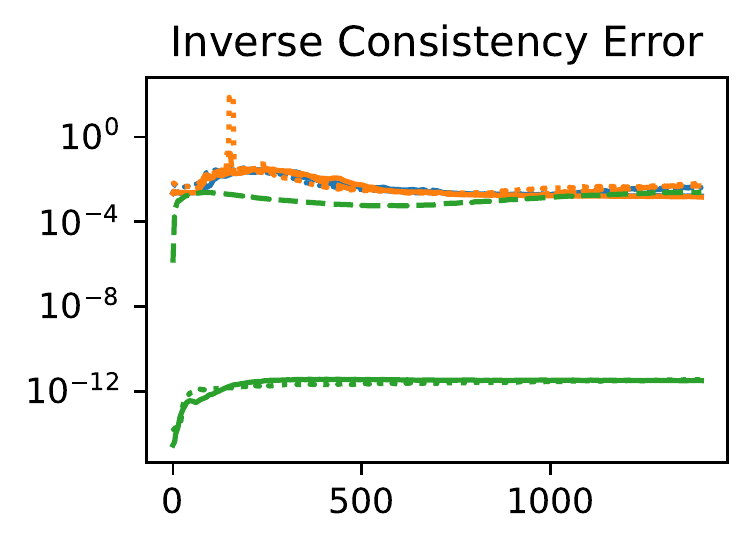}\hfill
	\includegraphics[width=0.32\textwidth]{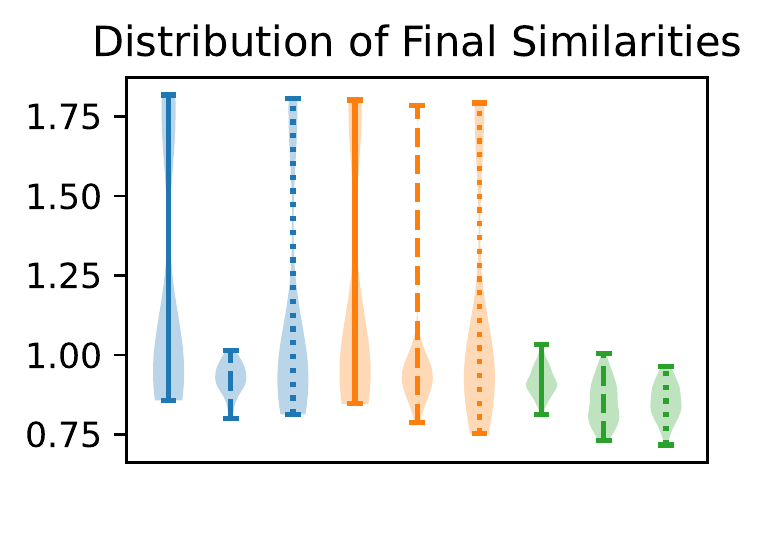}\\
	\vspace{-.6em}
	\includegraphics[width=0.62\textwidth]{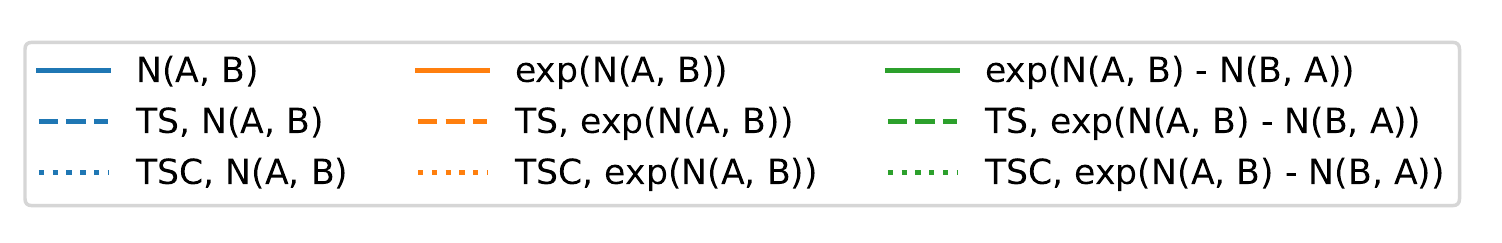}
	\vspace{-0.2cm}
	\caption{We vary the network used to perform affine registration and the method for composing steps on the Hollow Triangles and Circles dataset. Average loss curves and distribution of final losses are shown for 65 training runs per experimental configuration. Our \texttt{TwoStepConsistent} approach performs best overall. It shows fast convergence, high accuracy (indicated by a low similarity loss, left), is highly inverse consistent (middle), and trains reliably (indicated by the tight violin plot on the right).
	}
	\label{fig:affine_3x3}
\end{figure}

We observe that parameterizing an affine registration using the
$\exp(N_\theta^{AB} - N_\theta^{BA})$ construction speeds up the first few
epochs of training and gets even faster when combined with any multi-step
method. In Fig.~\ref{fig:affine_3x3}, note that in the top-left corner of the
first plot, the green loss curves (corresponding to models using
$N_\theta^{AB} - N_\theta^{BA}$) are roughly vertical, while the other loss
curves are roughly horizontal, eventually bending down. After this initial
lead, these green curves also converges to a better final loss. Further, all
methods that use the $N_\theta^{AB} - N_\theta^{BA}$ construction train
reliably, while other methods sometimes fail to converge
(Fig.~\ref{fig:affine_3x3}, right plot). This has a dramatic effect on the
practicality of a method since training on 3-D data can take multiple days on
expensive hardware.

Finally, as expected, the \emph{only} two approaches that are inverse consistent are the single-step inverse consistent by construction network, and the network using two inverse-consistent by construction subnetworks, joined by the
\texttt{TwoStepConsistent} operator. 
(Fig.~\ref{fig:affine_3x3}, middle, dotted and solid green).

\section{Evaluation on 3-D medical datasets}

We evaluate on several datasets, where we can compare to earlier registration
approaches. We use the network $\Phi := \texttt{TSC}\{\Psi_1,
	\texttt{TSC}\{\Psi_2, \texttt{TSC}\{\Xi_1, \Xi_2\}\}\}$ with $\Xi_i$
inverse-consistent SVF networks backed by U-Nets and $\Psi_i$
inverse-consistent affine networks backed by ConvNets\footnote{Specifically,
	$\texttt{networks.tallUNet2}$ and $\texttt{networks.ConvolutionalMatrixNet}$
	from the library icon\_registration version 1.1.1 on \texttt{pypi}}. We rely
on local normalized cross-correlation as our similarity measure, with $\sigma=5vx$, and regularize the SVF networks by the sum of the bending energies of their
velocity fields, with $\lambda=5$. We train end to end, minimizing
$-\text{LNCC}(I^A \circ \Phi[I^A, I^B], I^B) +
	\lambda\mathcal{L}_\text{reg}$ for 100,000 iterations ($\sim2$ days on 4 NVIDIA A6000s) with Adam
optimization and a learning rate of $1e$-$4$. In all cases, we normalize
images to the range (0, 1). We evaluate registration accuracy with and without
instance optimization \cite{wang2022plosl,lin2022GradICON}. Without instance
optimization, registration takes $\sim$0.23 seconds on an NVIDIA RTX A6000 on
the HCP ~\cite{van2012human} dataset. With instance optimization, registration
takes $\sim$43 seconds.

\subsection{Datasets}

\textbf{COPDGene / Dirlab lung CT.} We follow the data selection and preprocessing of \cite{lin2022GradICON}. We
train on 999 inhale/exhale pairs from COPDGene \cite{regan2011genetic}, masked
with lung segmentations, clipped to [-1000, 0] Hounsfield units, and scaled to
(0, 1). We evaluate landmark error (MTRE) on the ten inhale/exhale pairs
of the Dirlab challenge dataset~\cite{castillo2013reference}\footnote{\url{https://tinyurl.com/msk56ss5}}.

\vskip0.5ex
\noindent
\textbf{OAI Knee MRI}
We train and test on the split published with \cite{shen2019networks}, with
2532 training examples and 301 test pairs from the Osteoarthritis Initiative (OAI)~\cite{nevitt2006osteoarthritis}\footnote{\url{https://nda.nih.gov/oai}}. We evaluate using the mean Dice score of tibial and fibial cartilage. To compare directly to
\cite{shen2019networks,greer2021icon,lin2022GradICON} we train and evaluate at [80x192x192].

\vskip0.5ex
\noindent
\textbf{HCP Brain MRI}
We train on 1076 brain-extracted T1w images from the HCP
dataset \cite{van2012human} and test on a sample of 100 pairs between 36 images via mean Dice over 28 midbrain structures \cite{rushmore2022anatomically,rushmore_r_jarrett_2022_dataset}. We train and
execute the network at [130$\times$155$\times$130], then compute the Dice score at full resolution.

\vskip0.5ex
\noindent
\textbf{OASIS Brain MRI}
We use the OASIS-1~\cite{marcus2007open} data preprocessed by~\cite{hoopes2022learning}. This dataset contains images of 414 subjects. Following the data split in~\cite{Mok_2020_CVPR}, we train on 255 images and test on 153 images\footnote{Due to changes in the OASIS-1 data, our test set slightly differs from \cite{Mok_2020_CVPR}. We evaluate all methods using our testing protocol so that results are consistent}.
The images in the dataset are of size [160$\times$192$\times$224], and we crop the center of the image according to the preprocessing in~\cite{Mok_2020_CVPR}, leading to a size of [160$\times$144$\times$192]. During training, we sample image pairs randomly from the train set. For evaluation, we randomly pick 5 cases as the fixed images and register all the remaining 148 cases to the 5 cases, resulting in 740 image pairs overall.
\subsection{Comparisons}
We use publicly-available pretrained weights and code for
ANTs~\cite{avants2008symmetric}, PTVReg~\cite{vishnevskiy2017isotropic},
GradICON~\cite{lin2022GradICON}, SynthMorph~\cite{hoffmann2022synthmorph},
SymNet~\cite{Mok_2020_CVPR}, and EasyReg~\cite{iglesias2023easyreg}. SymNet,
GradICON, and PTVReg are run on the datasets associated with their original
publication. SynthMorph, which we evaluate on HCP, was originally trained and
evaluated on HCP-A and OASIS. EasyReg was trained on HCP~\cite{van2012human}
and ADNI~\cite{adni}. Our ConstrICON method outperforms the state of the art on HCP, OAI, and OASIS registration but underperforms on the DirLab data.
Since we use shared hyperparameters between these datasets, which are not
tuned to a specific task, we assert that this performance level will likely
generalize to new datasets. We find that our method is more inverse
consistent than existing inverse consistent by construction methods SymNet
 and SyNOnly with higher accuracy, and more inverse consistent than
inverse-consistent-by-penalty GradICON.

\begin{table}
	\caption{Results on 3-D medical registration. $\% |J|$ indicates the percentage of voxels with negative Jacobian. $\|\Phi^{AB} \circ \Phi^{BA} - \textrm{id}\|$ indicates the mean deviation in voxels from inverse consistency. Instance optimization is denoted by \emph{io}. Our ConstrICON approach shows excellent registration performance while being highly inverse consistent.}\label{tab1}
	\resizebox{1.\textwidth}{!}{
		\begin{tabular}{|l|l|l|l|l|l|l|l|l|}
			\cline{6-6} \cline{1-1}
			HCP                                           & \multicolumn{1}{c}{} & \multicolumn{1}{c}{} & \multicolumn{1}{c}{}                          &~~& DirLab                                 \\ \cline{6-9} \cline{1-4}
			Approach                                      & DICE                 & $ \% |J|$            & $\|\Phi^{AB} \circ \Phi^{BA} - \textrm{id}\|$ &  & Approach                              & MTRE & $ \% |J|$ & $\|\Phi^{AB} \circ \Phi^{BA} - id\|$          \\ \cline{1-4} \cline{6-9}
			ANTs SyNOnly~\cite{avants2008symmetric}                                   & 75.8                 & 0                    & 0.0350                                        &  & ANTs SyN~\cite{avants2008symmetric}    & 1.79 & 0         & 4.23                                          \\
			ANTs SyN                                      & 77.2                 & 0                    & 1.30                                          &  & PTVReg~\cite{vishnevskiy2017isotropic} & 0.96  & --        & 6.13                                            \\
			ConstrICON                                    & 79.3                 & 3.81e-6              & 0.000386                                      &  & ConstrICON                            & 2.03 & 6.6e-6    & 0.00518                                       \\
			ConstrICON + io                               & 80.1                 & 0                    & 0.00345                                       &  & ConstrICON + io                       & 1.62 & 3.02e-6   & 0.00306                                       \\
			GradICON~\cite{lin2022GradICON}                                     & 78.6                 & 0.00120              & 0.309                                         &  & GradICON~\cite{lin2022GradICON}        & 1.93 & 0.0003    & 0.575                                         \\
			GradICON + io                                 & 80.2                 & 0.000352             & 0.123                                         &  & GradICON + io                         & 1.31  & 0.0002    & 0.161                                         \\ \cline{6-9} 
			SynthMorph~\cite{hoffmann2022synthmorph} brain & 78.4                 & 0.364                & --                                            &  & OAI                                                                                                      \\ \cline{6-9}
			SynthMorph shape                              & 79.7                 & 0.298                & --                                            &  & Approach                              & DICE & $ \% |J|$ & $\|\Phi^{AB} \circ \Phi^{BA} - \textrm{id}\|$ \\ \cline{6-9} \cline{1-4}
			OASIS                                         & \multicolumn{1}{c}{} & \multicolumn{1}{c}{} & \multicolumn{1}{c}{}                          &  & ANTs SyN                              & 65.7 & 0         & 5.32                                          \\ \cline{1-4}
			Approach                                      & DICE                 & $ \% |J|$            & $\|\Phi^{AB} \circ \Phi^{BA} - \textrm{id}\|$ &  & GradICON                              & 70.4 & 0.0261    & 1.84                                          \\ \cline{1-4}
			ConstrICON                                    & 79.7                 & 9.73e-5              & 0.00776                                       &  & GradICON + io                         & 71.2 & 0.0042    & 0.504                                         \\ 
			SymNet~\cite{Mok_2020_CVPR}                   & 79.1                 & 0.00487              & 0.0595                                        &  & ConstrICON                            & 70.7 & 2.41e-7   & 0.0459                                        \\
			EasyReg~\cite{iglesias2023easyreg}            & 77.2                 & --                   & 0.181                                         &  & ConstrICON + io                       & 71.5 & 3.36e-7   & 0.0505                                        \\ \cline{1-4} \cline{6-9} 
		\end{tabular}
	}
\end{table}
\section{Conclusion}
The fundamental impact of this work is as a recipe for constructing a broad
class of exactly inverse consistent, multi-step registration algorithms. We
also are pleased to present registration results on four medically relevant
datasets that are competitive with the current state of the art, and in
particular are more accurate than existing inverse-consistent-by-construction
neural approaches.

\section{Acknowledgements}

This research was supported by NIH awards RF1MH126732, 1R01-AR072013, 1R01-HL149877, 1R01 EB028283, R41-MH118845, R01MH112748, 5R21LM013670, R01NS125307; by the Austrian Science Fund: FWF P31799-N38; and by the Land Salzburg (WISS 2025): 20102-F1901166-KZP, 20204-WISS/225/197-2019. The work expresses the views of the authors and not of the funding agencies. The authors have no conflicts of interest.


\bibliographystyle{splncs04}
\bibliography{references}
\newpage
\begin{figure}
	\resizebox{1.01\textwidth}{!}{
    \hspace*{-.5em}\begin{tabularx}{1.45\linewidth}{X X X X X X X X}
    Moving Image & Warped Image & $\Phi^{AB} \circ \Phi^{BA}$ & Fixed Image & Moving Image & Warped Image & $\Phi^{AB} \circ \Phi^{BA}$ & Fixed Image
    \end{tabularx}
    }
    \\
	\centering{
		\includegraphics[width=.118\textwidth]{Figures/medical_figures-2/Dirlab2Plane3image_A.png}
		\includegraphics[width=.118\textwidth]{Figures/medical_figures-2/Dirlab2Plane3warped_image_A_grid.png}
		\includegraphics[width=.118\textwidth]{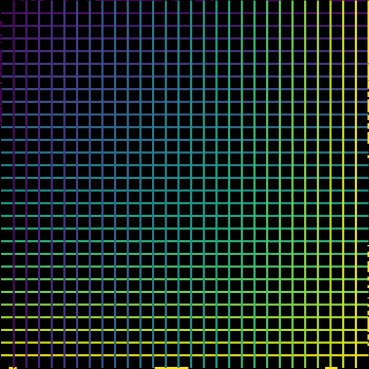}
		\includegraphics[width=.118\textwidth]{Figures/medical_figures-2/Dirlab2Plane3image_B.png}
		\includegraphics[width=.118\textwidth]{Figures/medical_figures-2/OAI2Plane1image_A.png}
		\includegraphics[width=.118\textwidth]{Figures/medical_figures-2/OAI2Plane1warped_image_A_grid.png}
		\includegraphics[width=.118\textwidth]{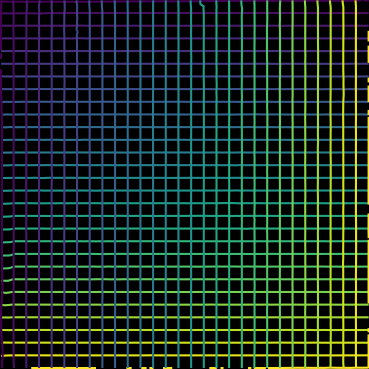}
		\includegraphics[width=.118\textwidth]{Figures/medical_figures-2/OAI2Plane1image_B.png}
	 	\\
		\includegraphics[width=.118\textwidth]{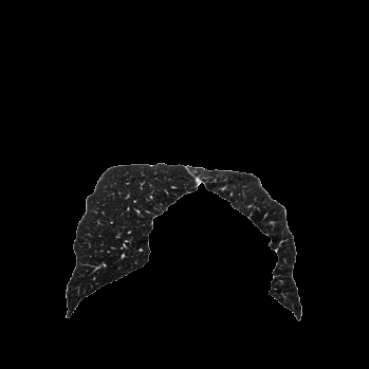}
		\includegraphics[width=.118\textwidth]{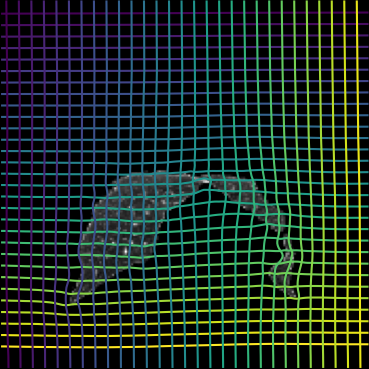}
		\includegraphics[width=.118\textwidth]{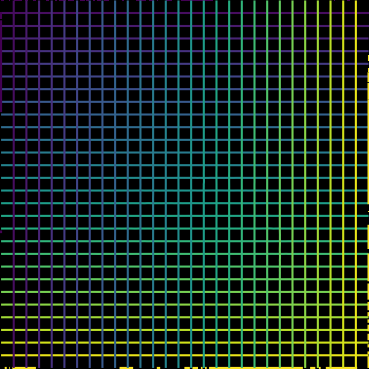}
		\includegraphics[width=.118\textwidth]{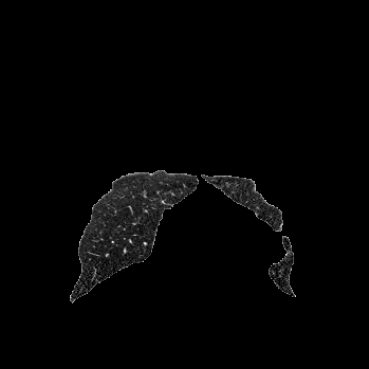}
		\includegraphics[width=.118\textwidth]{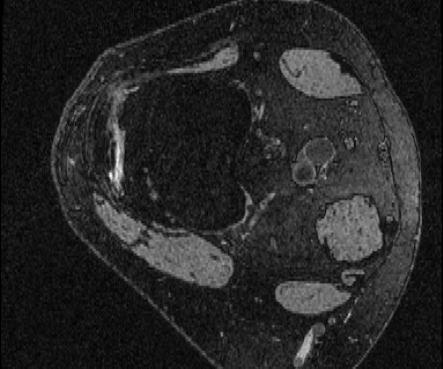}
		\includegraphics[width=.118\textwidth]{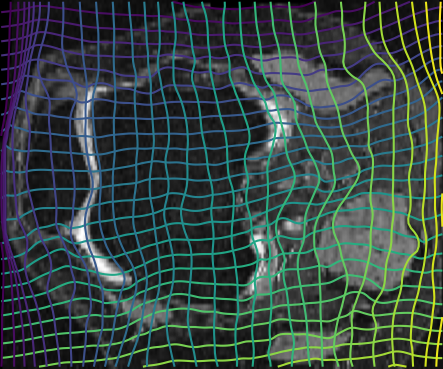}
		\includegraphics[width=.118\textwidth]{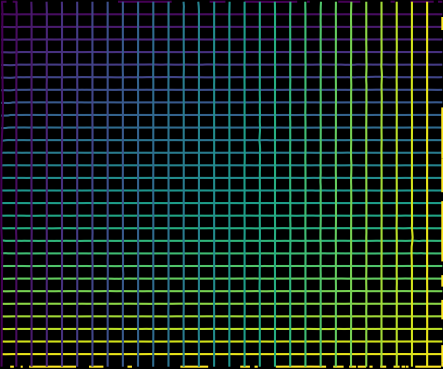}
		\includegraphics[width=.118\textwidth]{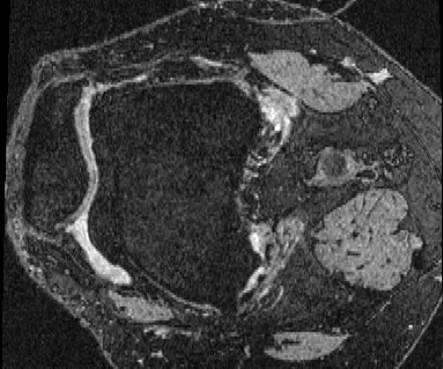}
        \\
		\includegraphics[width=.118\textwidth]{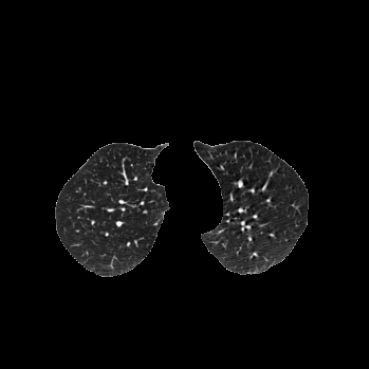}
		\includegraphics[width=.118\textwidth]{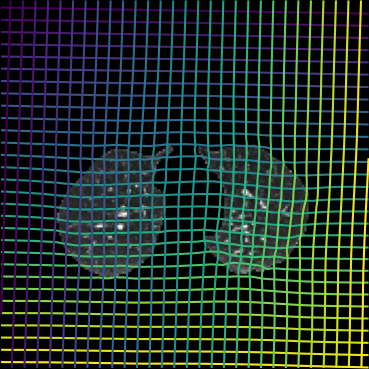}
		\includegraphics[width=.118\textwidth]{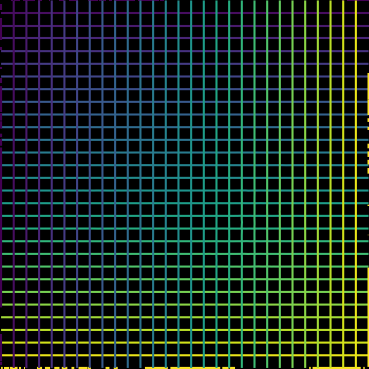}
		\includegraphics[width=.118\textwidth]{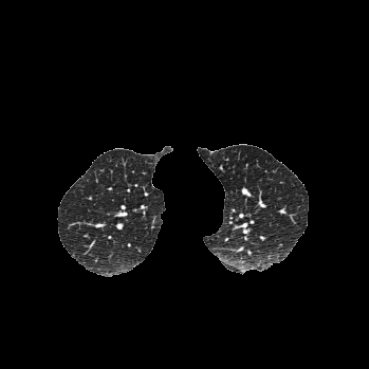}
		\includegraphics[width=.118\textwidth]{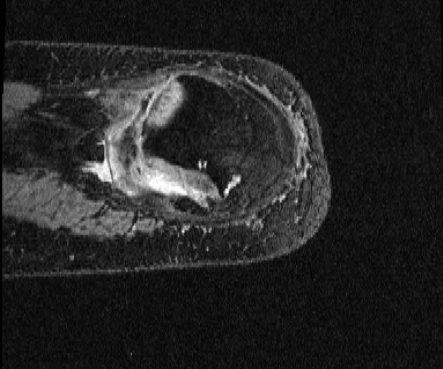}
		\includegraphics[width=.118\textwidth]{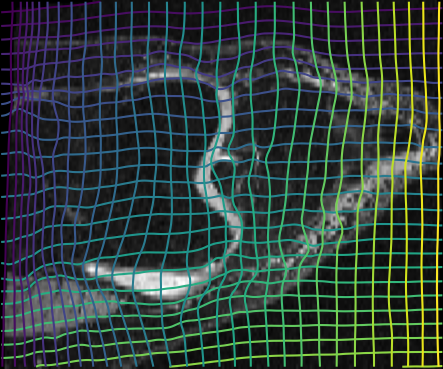}
		\includegraphics[width=.118\textwidth]{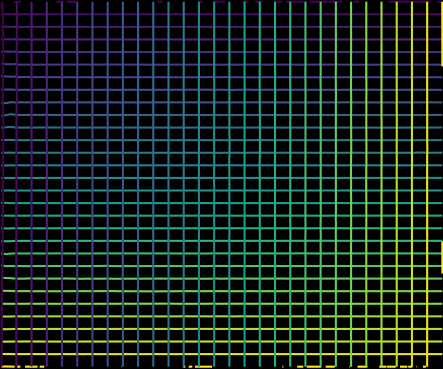}
		\includegraphics[width=.118\textwidth]{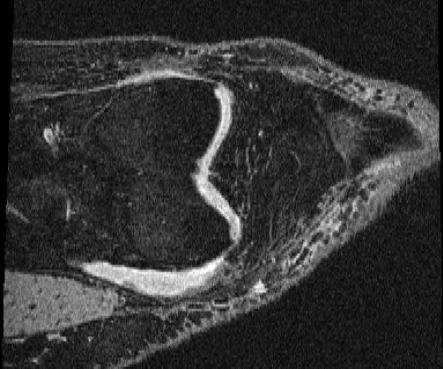}
        \\
		\includegraphics[width=.118\textwidth]{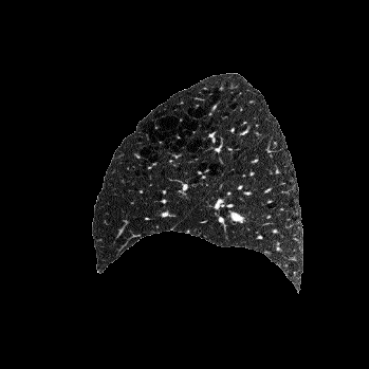}
		\includegraphics[width=.118\textwidth]{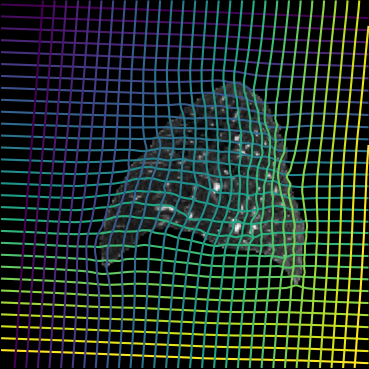}
		\includegraphics[width=.118\textwidth]{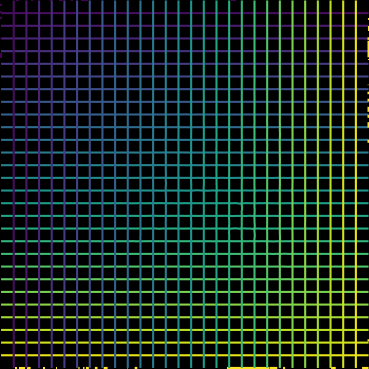}
		\includegraphics[width=.118\textwidth]{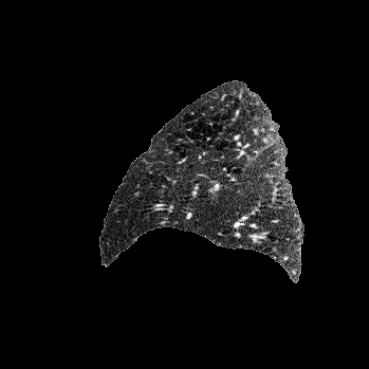}
		\includegraphics[width=.118\textwidth]{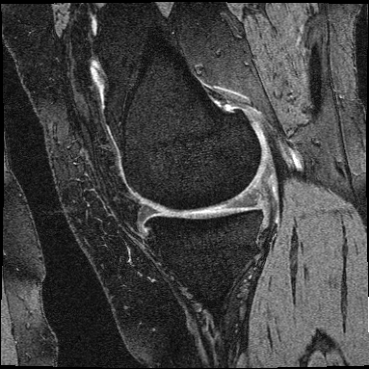}
		\includegraphics[width=.118\textwidth]{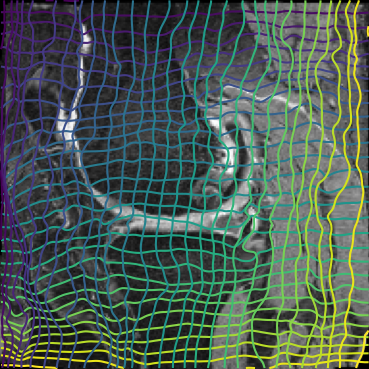}
		\includegraphics[width=.118\textwidth]{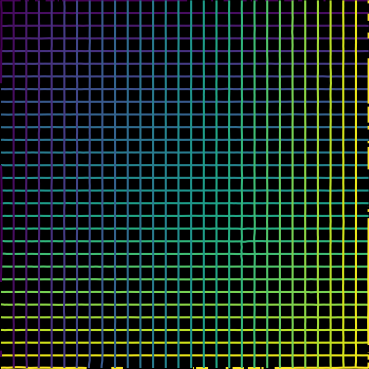}
		\includegraphics[width=.118\textwidth]{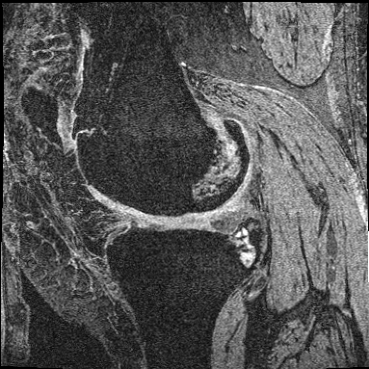}
	 	\\
		\includegraphics[width=.118\textwidth]{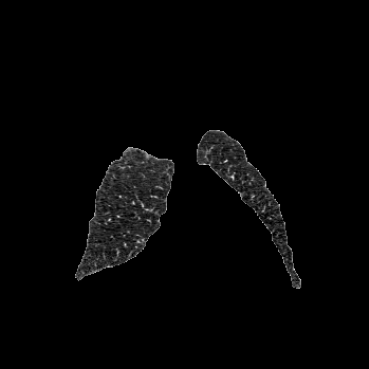}
		\includegraphics[width=.118\textwidth]{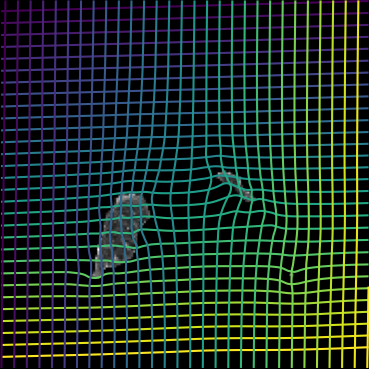}
		\includegraphics[width=.118\textwidth]{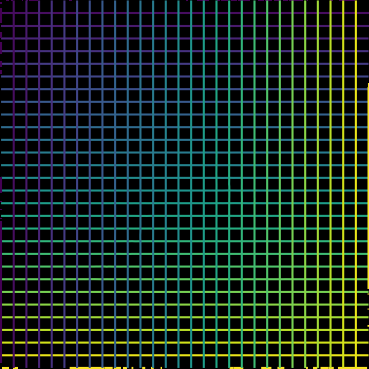}
		\includegraphics[width=.118\textwidth]{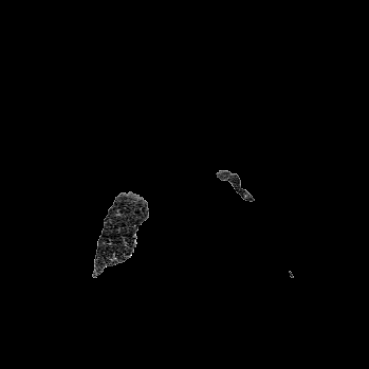}
		\includegraphics[width=.118\textwidth]{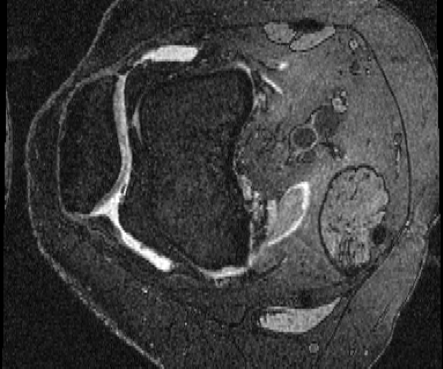}
		\includegraphics[width=.118\textwidth]{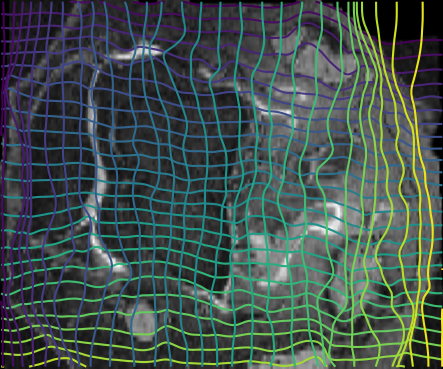}
		\includegraphics[width=.118\textwidth]{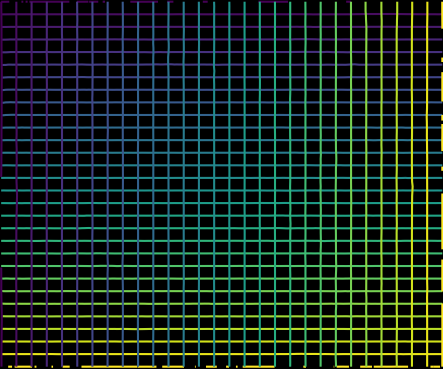}
		\includegraphics[width=.118\textwidth]{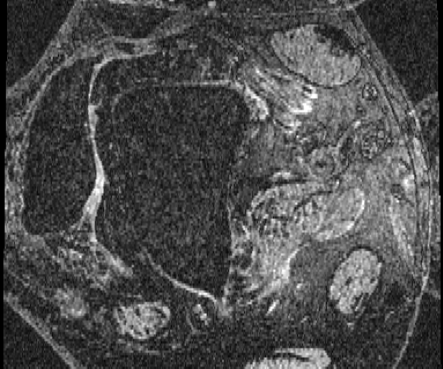}
        \\
		\includegraphics[width=.118\textwidth]{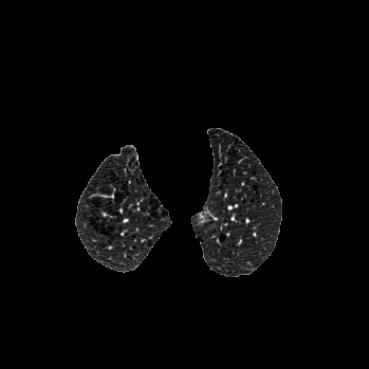}
		\includegraphics[width=.118\textwidth]{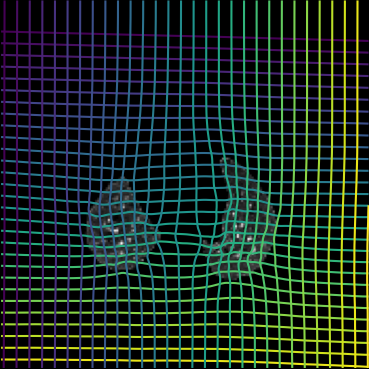}
		\includegraphics[width=.118\textwidth]{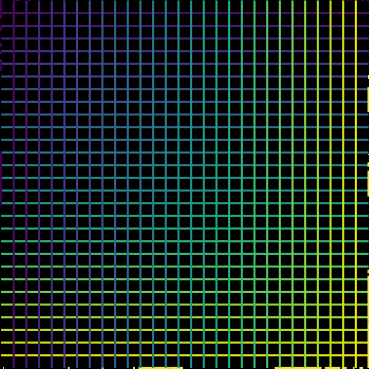}
		\includegraphics[width=.118\textwidth]{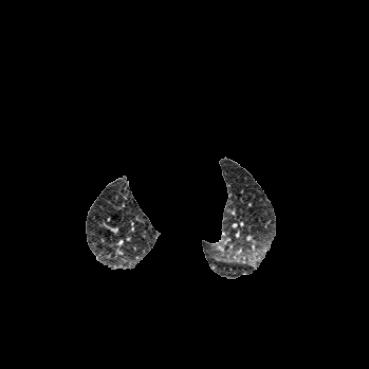}
		\includegraphics[width=.118\textwidth]{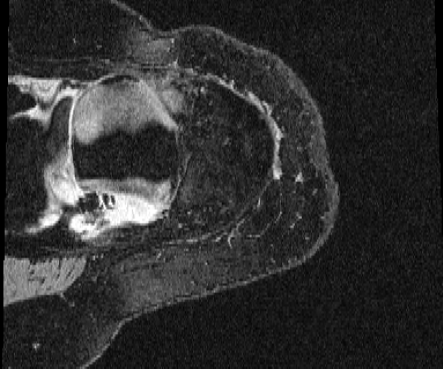}
		\includegraphics[width=.118\textwidth]{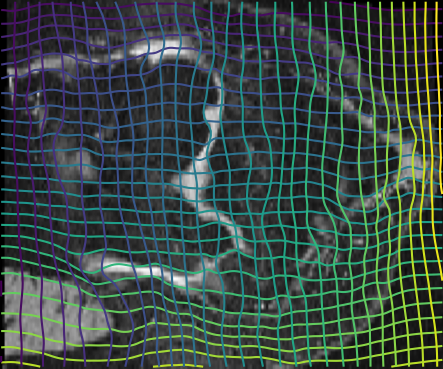}
		\includegraphics[width=.118\textwidth]{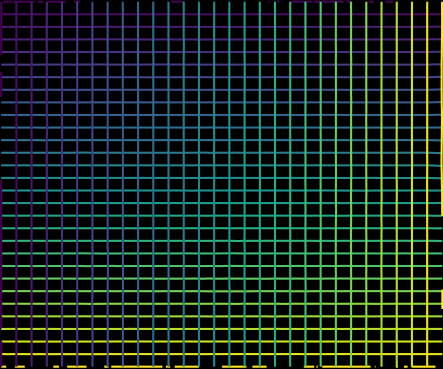}
		\includegraphics[width=.118\textwidth]{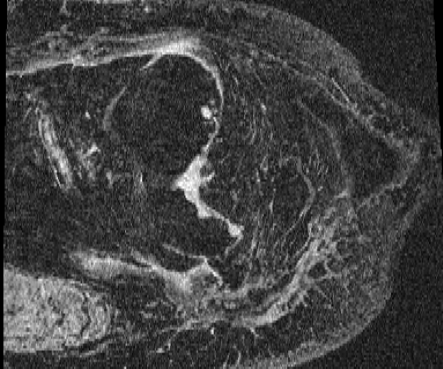}
        \\
	 	\includegraphics[width=.118\textwidth]{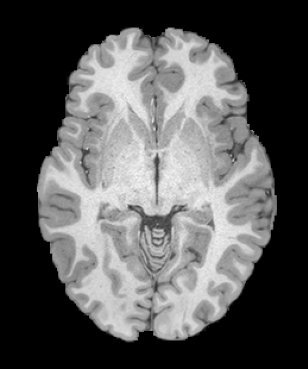}
	 	\includegraphics[width=.118\textwidth]{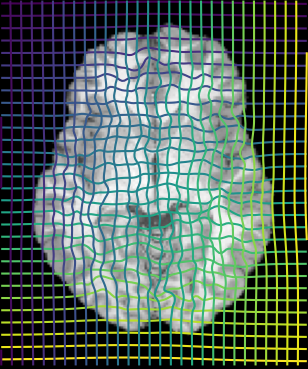}
	 	\includegraphics[width=.118\textwidth]{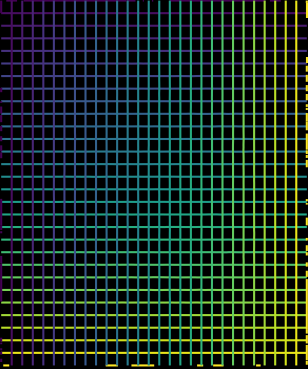}
	 	\includegraphics[width=.118\textwidth]{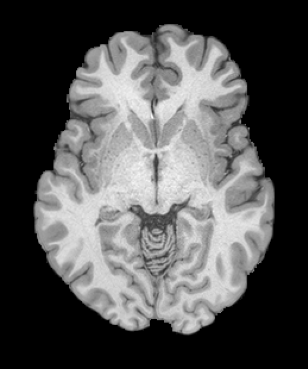}
	 	\includegraphics[angle=180,origin=c,width=.118\textwidth]{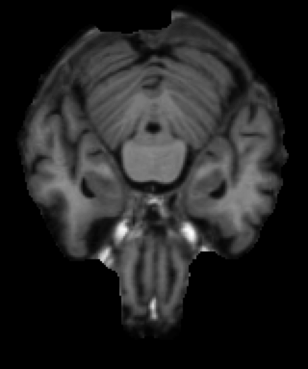}
	 	\includegraphics[angle=180,origin=c,width=.118\textwidth]{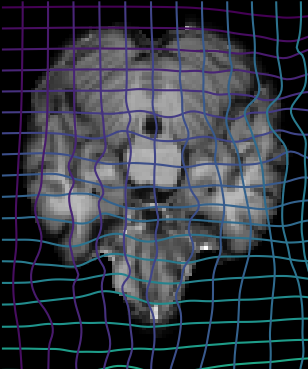}
	 	\includegraphics[angle=180,origin=c,width=.118\textwidth]{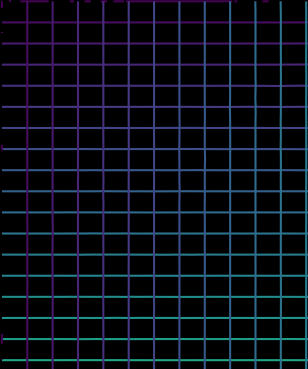}
	 	\includegraphics[angle=180,origin=c,width=.118\textwidth]{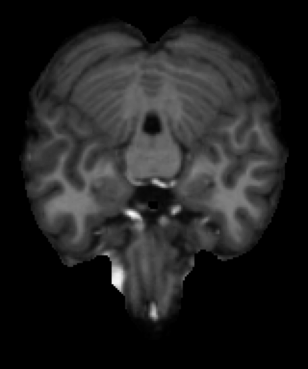}
        \\
	 	\includegraphics[width=.118\textwidth]{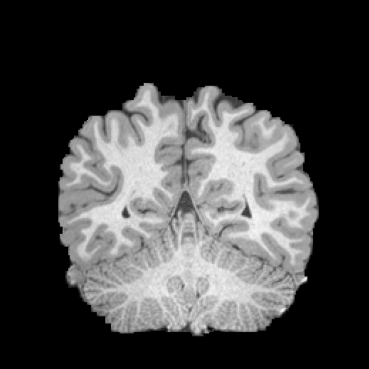}
	 	\includegraphics[width=.118\textwidth]{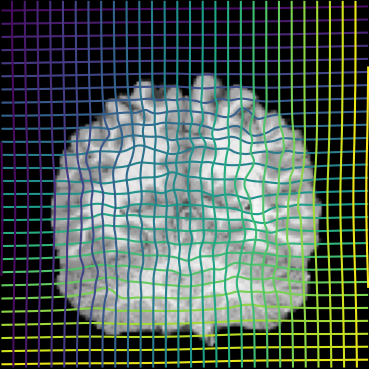}
	 	\includegraphics[width=.118\textwidth]{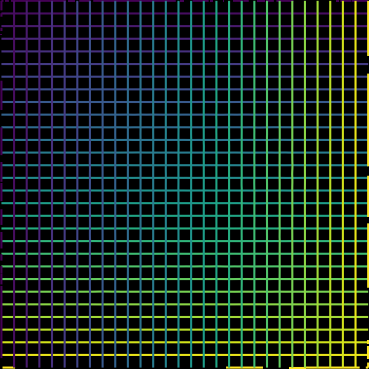}
	 	\includegraphics[width=.118\textwidth]{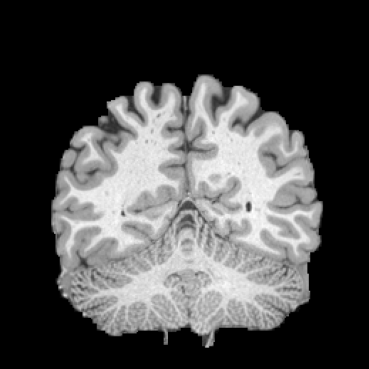}
	 	\includegraphics[width=.118\textwidth]{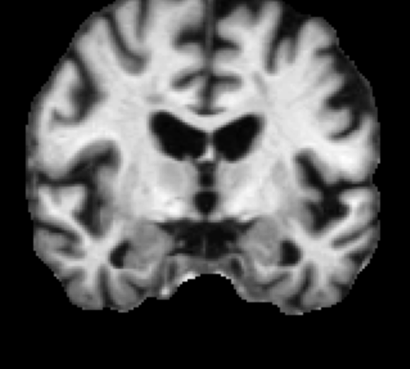}
	 	\includegraphics[width=.118\textwidth]{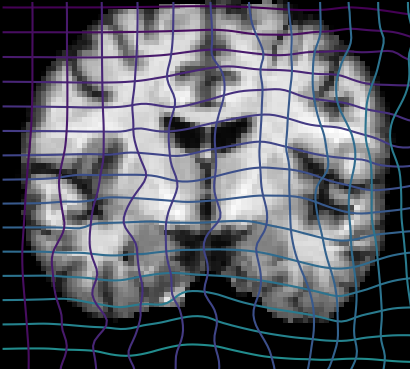}
	 	\includegraphics[width=.118\textwidth]{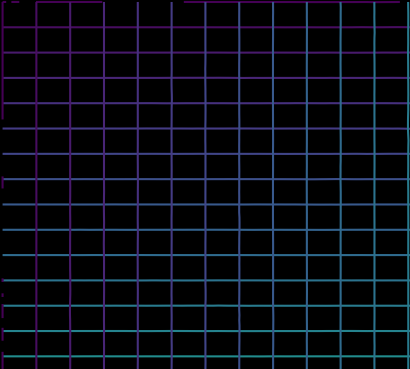}
	 	\includegraphics[width=.118\textwidth]{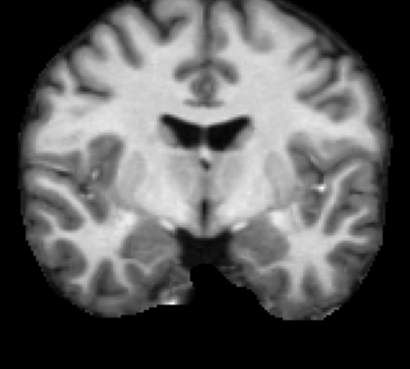}
        \\
	 	\reflectbox{\includegraphics[width=.118\textwidth]{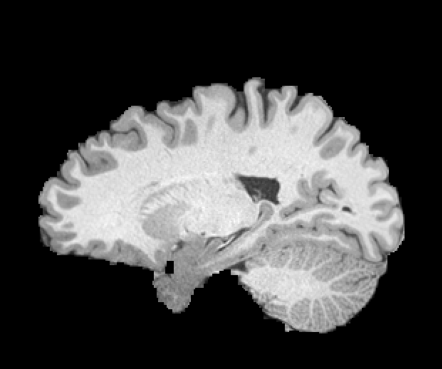}}
	 	\reflectbox{\includegraphics[width=.118\textwidth]{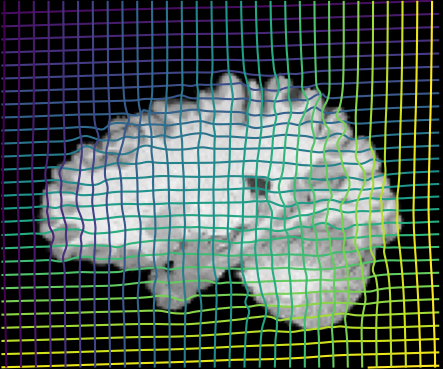}}
	 	\reflectbox{\includegraphics[width=.118\textwidth]{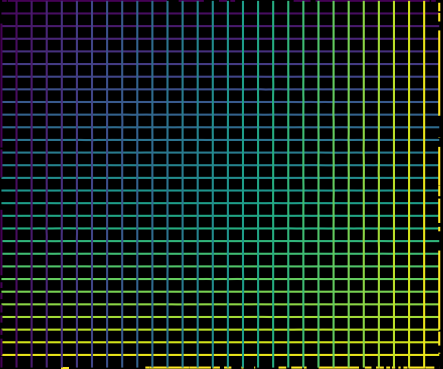}}
	 	\reflectbox{\includegraphics[width=.118\textwidth]{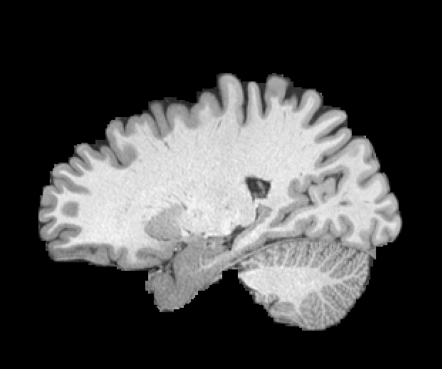}}
	 	\reflectbox{\includegraphics[angle=270,origin=rB,width=.118\textwidth]{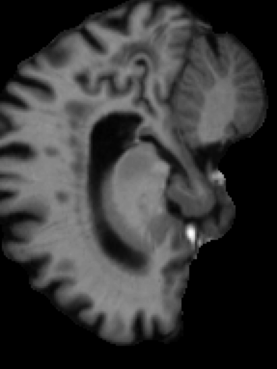}}
	 	\reflectbox{\includegraphics[angle=270,origin=rB,width=.118\textwidth]{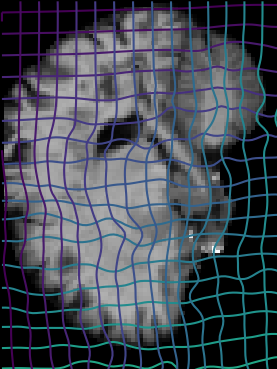}}
	 	\reflectbox{\includegraphics[angle=270,origin=rB,width=.118\textwidth]{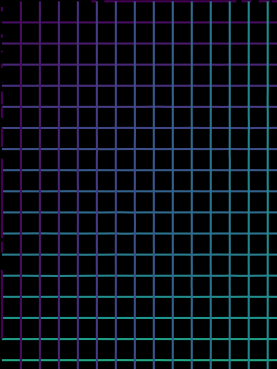}}
	 	\reflectbox{\includegraphics[angle=270,origin=rB,width=.118\textwidth]{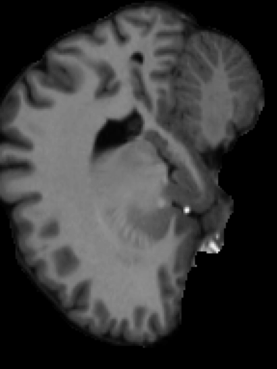}}
        \\
	 	\includegraphics[width=.118\textwidth]{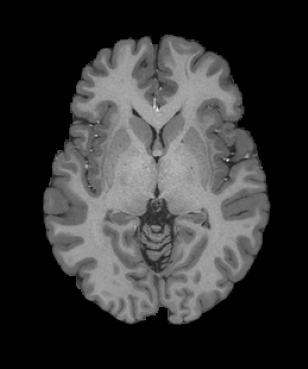}
	 	\includegraphics[width=.118\textwidth]{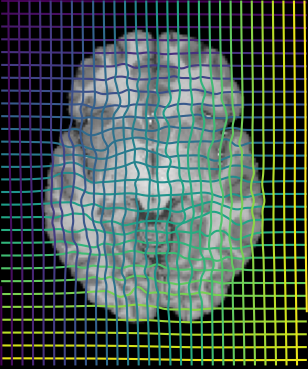}
	 	\includegraphics[width=.118\textwidth]{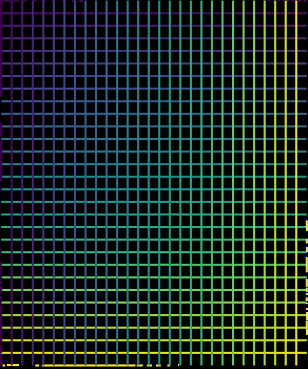}
	 	\includegraphics[width=.118\textwidth]{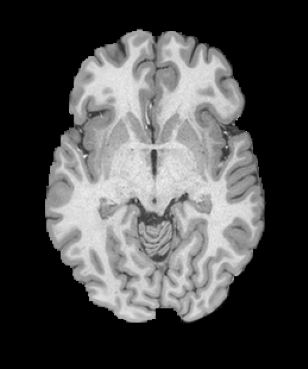}
	 	\includegraphics[angle=180,origin=c,width=.118\textwidth]{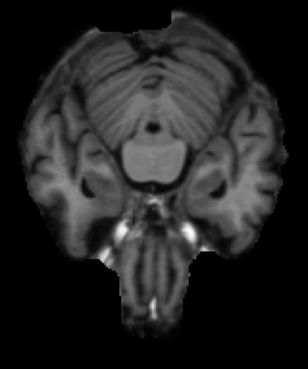}
	 	\includegraphics[angle=180,origin=c,width=.118\textwidth]{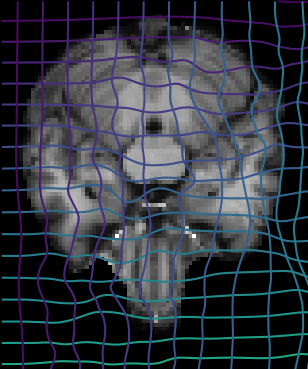}
	 	\includegraphics[angle=180,origin=c,width=.118\textwidth]{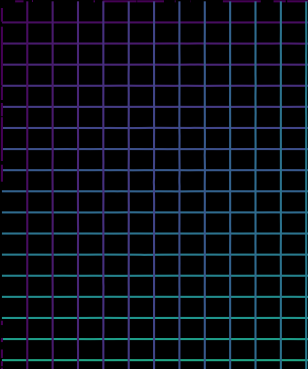}
	 	\includegraphics[angle=180,origin=c,width=.118\textwidth]{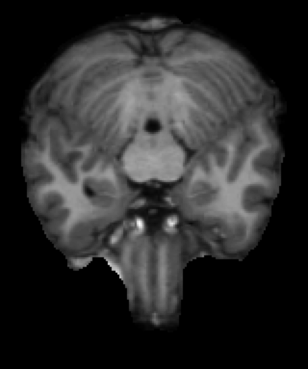}
        \\
	 	\includegraphics[width=.118\textwidth]{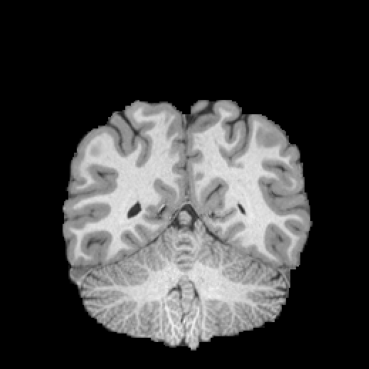}
	 	\includegraphics[width=.118\textwidth]{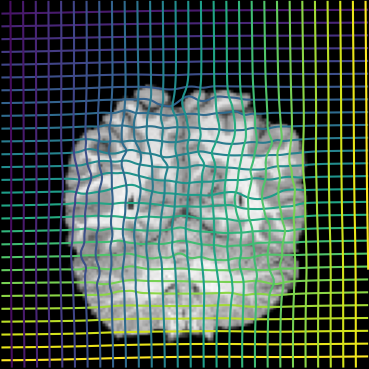}
	 	\includegraphics[width=.118\textwidth]{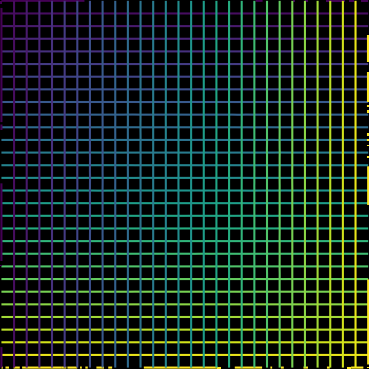}
	 	\includegraphics[width=.118\textwidth]{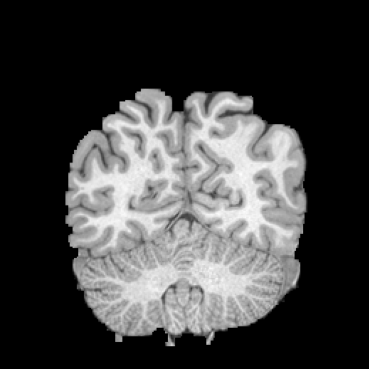}
	 	\includegraphics[width=.118\textwidth]{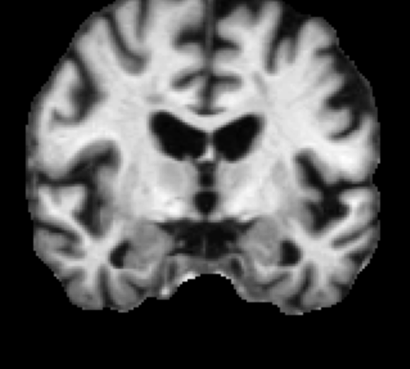}
	 	\includegraphics[width=.118\textwidth]{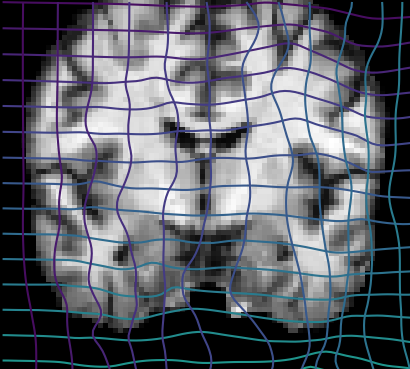}
	 	\includegraphics[width=.118\textwidth]{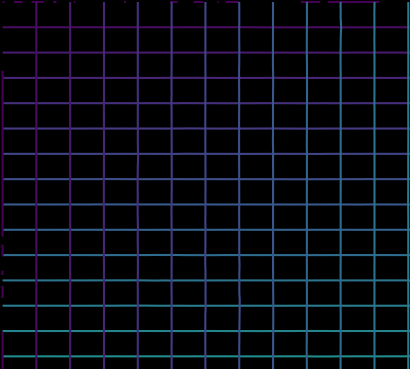}
	 	\includegraphics[width=.118\textwidth]{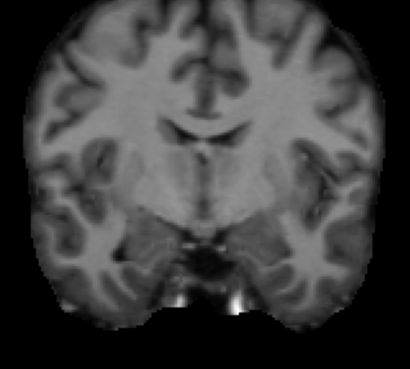}
        \\
	 	\reflectbox{\includegraphics[width=.118\textwidth]{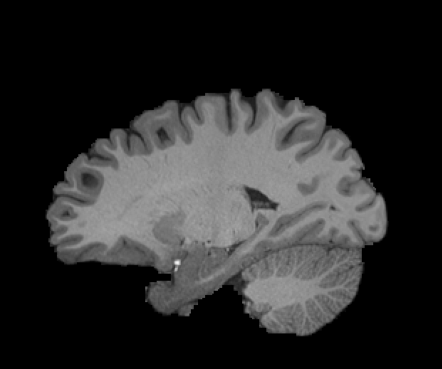}}
	 	\reflectbox{\includegraphics[width=.118\textwidth]{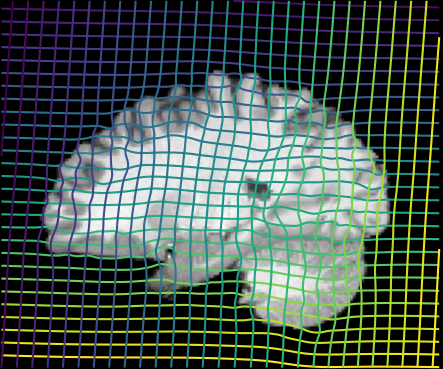}}
	 	\reflectbox{\includegraphics[width=.118\textwidth]{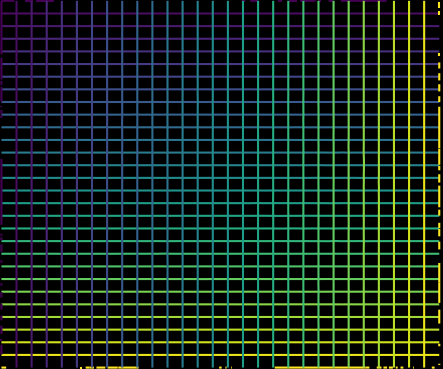}}
	 	\reflectbox{\includegraphics[width=.118\textwidth]{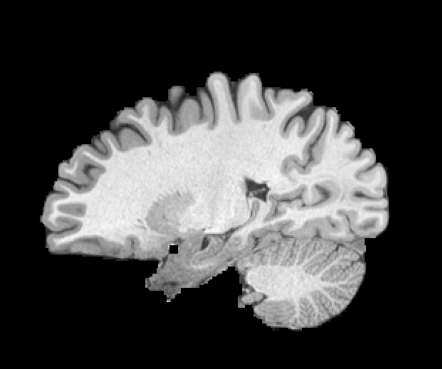}}
	 	\reflectbox{\includegraphics[angle=270,origin=rB,width=.118\textwidth]{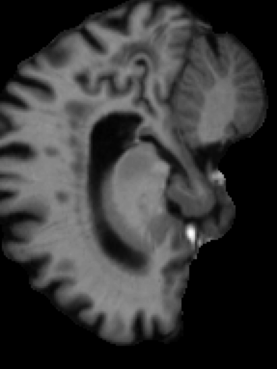}}
	 	\reflectbox{\includegraphics[angle=270,origin=rB,width=.118\textwidth]{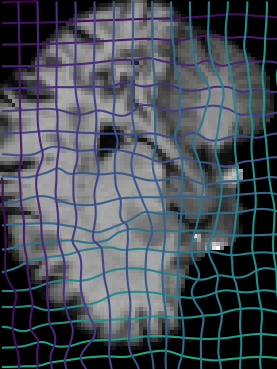}}
	 	\reflectbox{\includegraphics[angle=270,origin=rB,width=.118\textwidth]{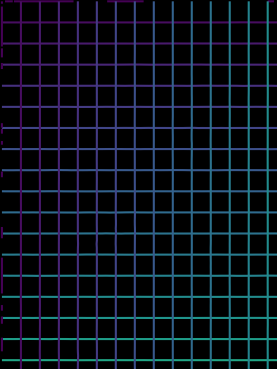}}
	 	\reflectbox{\includegraphics[angle=270,origin=rB,width=.118\textwidth]{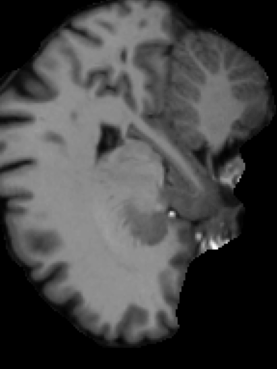}}
        \\
   }
	\caption{Cases registered by \texttt{ConstrICON} from DirLab (top left) , OAI (top right), HCP (bottom left), and OASIS (bottom right). For each test set two pairs are shown, each sliced along three axes. The maps generated by composing $\Phi^{AB} \circ \Phi^{BA}$ are almost indistinguishable from identity.}
	\label{fig:medical_sample}
	\vspace{-2em}
\end{figure}

\begin{figure}
    \centering
    \includegraphics[width=.5\textwidth]{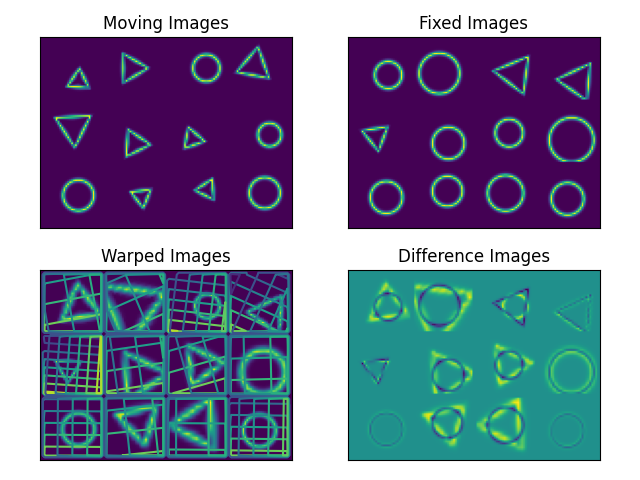}
    \caption{For brevity, in the main paper visualizations of the registrations performed in Fig. 3 were omitted. Here, we include sample affine registrations of pairs from the Hollow Triangles and Circles  dataset in that experiment, performed by the model composed of two inverse consistent networks joined by the \texttt{TwoStepConsistent} operator (denoted $\text{TSC},~\exp(N^{AB} - N^{BA})$ in Figure 3)}
    \label{fig:my_label}
\end{figure}

\begin{figure}
    \centering
    \includegraphics[width=1\textwidth]{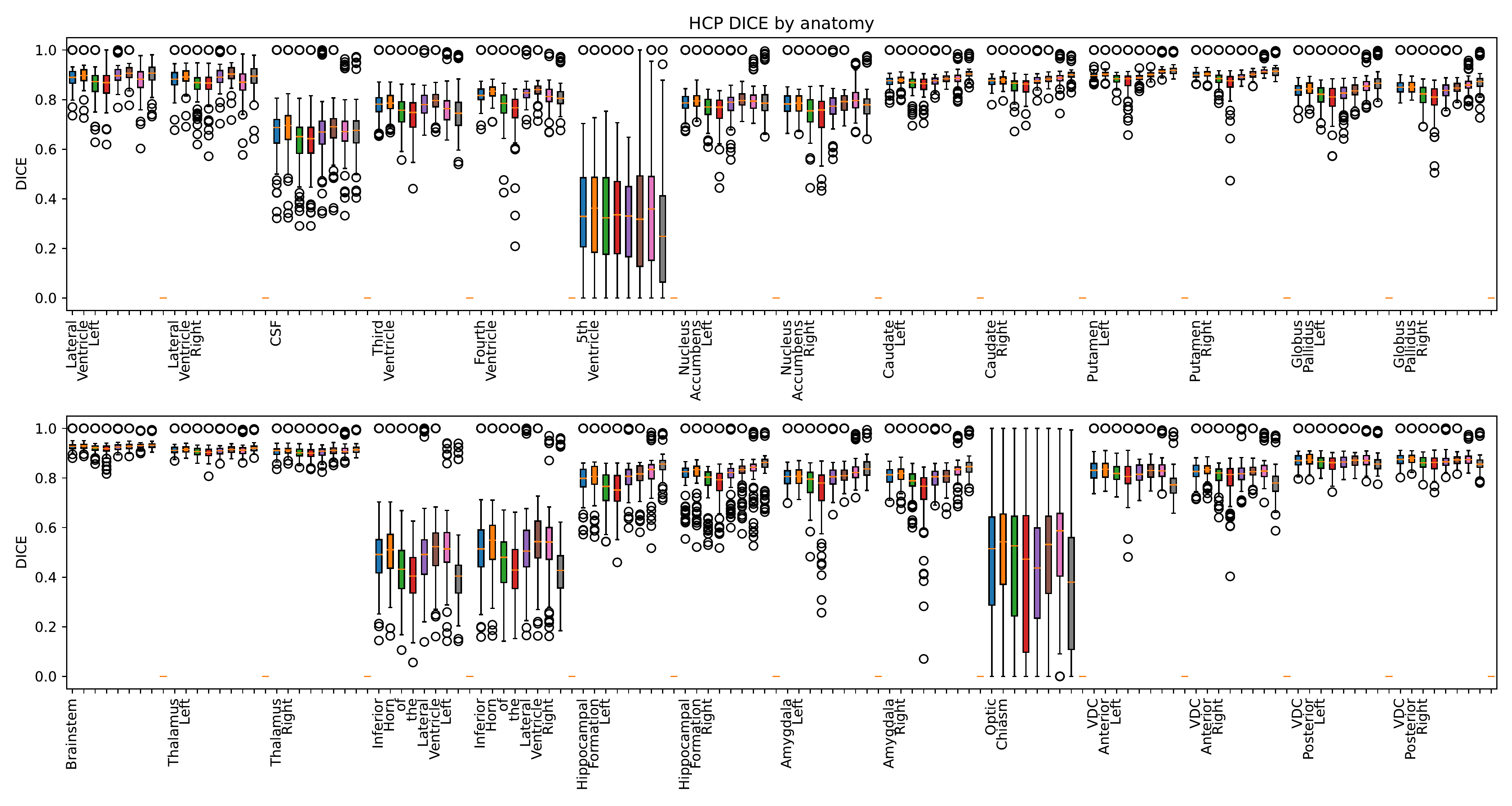}\\
    \includegraphics[width=.5\textwidth]{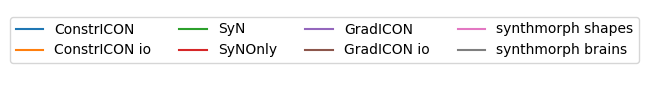}\\

    \caption{DICE scores on the HCP test set, broken down by registration approach and segmented anatomy. Note that as the test set was selected by drawing random pairs from a set of 36 images, it contains a few instances of registering an image to itself. (This inclusion matches the evaluation on HCP in ~\cite{lin2022GradICON}) These instances demonstrate an interesting effect: by definition, the inverse consistent methods (ConstrICON, ANTs SyNOnly) score perfectly on these pairs, creating the outliers seen at DICE 1. The approximately inverse consistent methods (GradICON, ANTs SyN) score close to perfectly.  Synthmorph scores better on these pairs than on non-identical pairs, but still generates some deformation, reducing the score from 1.}
    \label{fig:brain_violin}
\end{figure}

\end{document}